\documentclass{bmvc2k}

\title{\Large{Functionality-Oriented Convolutional Filter Pruning}}
\addauthor{Zhuwei Qin}{zqin@gmu.edu}{1}
\addauthor{Fuxun Yu}{fyu2@gmu.edu}{1}
\addauthor{Chenchen Liu}{ccliu@umbc.edu}{2}
\addauthor{Xiang Chen}{xchen26@gmu.edu}{1}

\addinstitution{
 School of Electrical and Computer Engineering \\
 George Mason University \\
 Fairfax, VA, USA
}
\addinstitution{
 University of Maryland, \\ Baltimore County \\  
 Baltimore, MD, USA
}

\runninghead{QIN, ET AL}{Functionality-Oriented Convolutional Filter Pruning}

\usepackage{comment}
\usepackage{color}
\usepackage{amsmath}
\usepackage{algorithm}
\usepackage{algorithmicx}	
\usepackage{varwidth}
\usepackage{algpseudocode}
\usepackage{wrapfig}
\usepackage{graphicx}
\usepackage{colortbl} 
\usepackage{arydshln}
\usepackage{threeparttable}

\graphicspath{{_fig/}}

\DeclareMathOperator*{\argmax}{argmax}

\usepackage{tabularx, booktabs}
\newcolumntype{I}{!{\vrule width 3pt}}
\newlength\savedwidth

\newlength\savewidth

\algtext*{EndWhile}
\algtext*{EndIf}
\algtext*{EndFor}
\algtext*{EndProcedure}

\begin{document}

\maketitle
\begin{abstract}
The sophisticated structure of Convolutional Neural Network (CNN) models allows for outstanding performance, but at the cost of intensive computation load.
	To reduce this cost, many model compression works have been proposed to eliminate insignificant model structures, such as pruning the convolutional filters which have smaller absolute weights.
	However, most of these works merely depend on quantitative significance ranking without qualitative filter functionality interpretation or thorough model structure analysis, resulting in considerable model retraining cost.
Different from previous works, we interpret the functionalities of the convolutional filters and identify the model structural redundancy as repetitive filters with similar feature preferences.
	In this paper, we proposed a functionality-oriented filter pruning method, which can precisely remove the redundant filters without compromising the model functionality integrity and accuracy performance.
Experiments with multiple CNN models and databases testified the unreliability of conventional weight-ranking based filter pruning methods, and demonstrate our method's advantages in terms of computation load reduction (at most 68.88\% FLOPs), accuracy retaining ($\textless$0.34\% accuracy drop), and expected retraining independence.

\end{abstract}

\section{Introduction}
\begin{figure}[t]\label{fig:norm}
	\centering
	\includegraphics[width=5in]{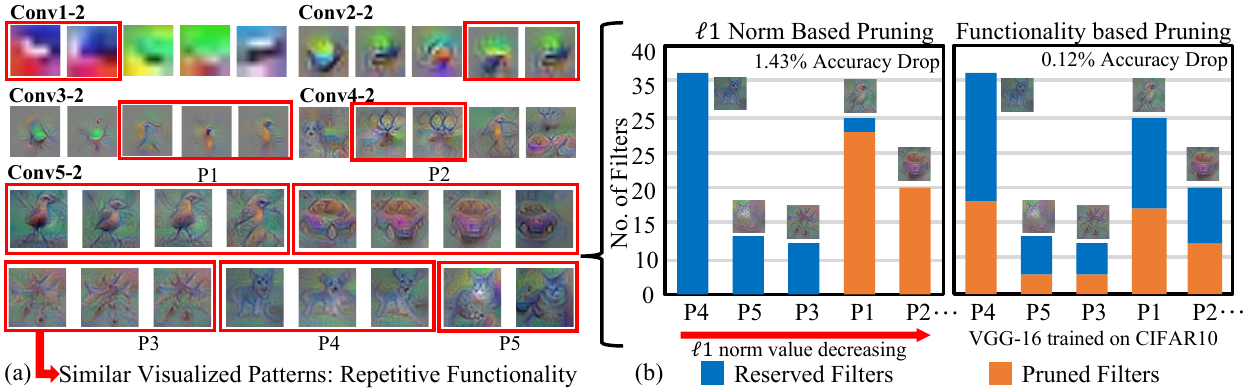}
	\vspace{-2mm}
	\caption{(a) An illustration of the visualized patterns of convolutional filters. (b) $\ell_1$ norm based filter pruning and functionality-oriented pruning.}
	\vspace{-3mm}
\end{figure}
Nowadays, Convolutional Neural Networks (CNNs) have been widely applied to various cognitive tasks (\textit{esp.} image recognition)~\cite{Kriz2012NIPS,luo2018taking,zhu2019sim}.
	This great success is benefited from CNNs' sophisticated model structures, which utilize multiple inter-connected layers of convolutional filters to hierarchically abstract input data features and assemble accurate prediction results~\cite{Gonzalez:2018:semantic,He:2016:CVPR}.
	However, the increasing layer width and depth boost not only accuracy but also computation load.
	For example, a 2.3$\times$ parameter size increment from AlexNet to VGG-16 will introduce 21.4$\times$ more computation load in terms of FLOPs~\cite{Kriz2012NIPS,Simo:2014:arXiv}.

	Many optimization works have been proposed to relieve the CNN computation cost~\cite{Han:2015:deep,jaderberg2014speeding,Li:2016:pruning}, and the filter pruning is considered as one of the most efficient approaches.
	Filter pruning methods identify the convolutional filters with the smallest significance based on different metrics.
	By removing these filters and repeatedly retraining the model, the computation load can be significantly reduced while the prediction accuracy is well retained~\cite{he2018GM,zhuo2018scsp,yu2018nisp}.
	However, such significance-ranking based filter pruning methods have been increasingly questioned:
	On the one hand, the correlation between the filter weight and functionality significance (\textit{e.g.}, in~\cite{Li:2016:pruning}) is not theoretically proven. Some works have found that pruning certain small filters may cause severe accuracy drop than pruning large ones~\cite{ye2018rethinking}.
	On the other hand, the excessive dependence on the retraining process seriously questions the rationality of the conventional filter significance identification.
	Some works show that the retraining process actually reconstructs the CNN models due to massive filters are inappropriately pruned~\cite{liu2018rethinking}.
	Therefore, to re-examine the convolutional filter pruning, rather than merely quantitatively ranking the filter significance, it is urgent to qualitatively interpret filter functionalities and identify actual model redundancies.

In this work, targeting image-cognitive CNNs, we utilized CNN visualization techniques to interpret the convolutional filter functionality~\cite{Zhou:2016:CVPR:Network-Dissection,zhou2014detectors,Yosinski:2015:ICML:AM}.
	Specifically, we adopted the Activation Maximization method to synthesize a specific input image for each filter~\cite{Yosinski:2015:ICML:AM}.
	These images can trigger the maximum activation of corresponding filters, thus representing each filter's preferred input feature pattern as its particularly functionality.
	As shown in Fig.~\ref{fig:norm}~(a), with the layer depth increasing, the visualized filter functionality patterns evolve from fundamental colors and shapes into recognizable objects.

Based on the visualized interpretation, we discover that a certain number of repetitive filters with similar functionality patterns exist in each layer, which can be grouped into multiple functionality clusters (denoted by red boxes).
	The functionality repetition in each cluster can be considered as a model structural redundancy in the filter level.

Motivated by this discovery, we propose a functionality-oriented filter pruning method.
	In our methods, regardless of the weight values, we find the best pruning ratio of every cluster in every layer to reduce the model structural redundancy. After the pruning, only a small amount of retraining effort is required to fine tune the accuracy performance.
	Fig.~\ref{fig:norm}~(b) presents a set of filter pruning examples to demonstrate the difference between our proposed method and the conventional $\ell_1$-norm based filter pruning method.
	With the same amount of filters pruned, we can see that, our method precisely address the redundant filters in every cluster and layer, resulting a negligible accuracy drop.

Experiment results show that, on CIFAR-10, our method reduces more than 43\% and 44.1\% FLOPs on ResNet-56 and VGG-16 respectively, achieving 0.05\% and 0.72\% relative accuracy improvement.  
On CIFAR100, our method reduces more than 37.2\% FLOPs on VGG-16 without losing accuracy.
On ImageNet, with 50.64\% and 23.8\% FLOPs reduction, our method can accelerate VGG-16 and ResNet32 without accuracy drop.


We also observe the filter functionality transition in the retraining process.
	Experiments reveal that, the conventional filter pruning methods may significantly compromise filters' functionality integrity, resulting in considerable retraining cost.
	While, our method demonstrates expected accuracy retaining capability and retraining independence.

In the following sections, we will proceed into specific design and evaluation details.


\section{Related work}
\label{sec:prelim}
\textbf{Convolutional Filter Pruning}
	Previous filter pruning works can be roughly divided into two categories:
%
%
(1) \textit{Post-Training Filter Pruning} methods are applied to pre-trained CNN models by identifying and pruning the insignificant filters based on particular weights ranking schemes.
	For example,~\cite{Li:2016:pruning},~\cite{he2018soft}, and~\cite{molchanov2016taylor} utilized $\ell_1$-norm, $\ell_2$-norm, and Taylor expansion respectively to rank the filter weights for pruning in each layer.
(2) \textit{Training Phase Filter Pruning} methods apply particular regulation constraints to the CNN model training phase, and enforce a certain amount of filters to become sufficiently small to be safely pruned (\textit{e.g.}, structured sparsity learning~\cite{wen2016learning}, structured Bayesian pruning~\cite{neklyudov2017bayesian}, and $\ell_0$ regularization~\cite{louizos2017learning}).
Usually, the second category methods can achieve better optimization performance due to their profound effects in the early training stage.
	However, the first category methods have better practicability with wider application scenarios.
	In this work, we will focus on renovating the first category methods and compare our methods with corresponding state-of-the-art.

\noindent\textbf{Convolutional Filter Visualization}
As the convolutional filters are designed to capture certain input features, the semantics of the captured feature can conclusively indicate the functionality of each filter~\cite{qin2018visualization,mahendran2015:Network-Inversion,zhou2018revisiting,qin2019captor}.
	However, the functionality is hard to be directly interpreted, which significantly hinders qualitative CNN model analysis and optimization development~\cite{Gonzalez:2018:semantic}.
	Recently, many CNN visualization works have been proposed to analyze CNN models in a functionality perspective:~\cite{Zhou:2016:CVPR:Network-Dissection} established the correlation between each filter and a specific semantic concept;~\cite{Yosinski:2015:ICML:AM} designed a novel visualization technique -- Activation Maximization -- to illustrated a filter's maximum activation pattern, which represents the filter's exclusive feature preference as its functionality.
In this work, we will utilize the Activation Maximization (AM) as our major visualization tool for filter analysis.
\section{Functionality-Oriented Filter Pruning Methods}
\label{sec:pruning}
\noindent\textbf{Proposed Method Overview}
Based on the convolutional filter functionality and structural redundancy analysis, we propose a functionality-oriented filter pruning method.
The proposed method consists of the following major steps:

\vspace{0.5mm}
(1) \textit{Filter Functionality Interpretation}: Given a pre-trained model, each filter's functionality is firstly interpreted by AM visualization;

\vspace{0.5mm}
(2) \textit{Functionality Redundancy Identification}: Based on proper similarity analysis on the visualized functionality patterns, the filters with repetitive functionalities are clustered together.
	These repetitive filters can be considered as redundant filters to be pruned;

\vspace{0.5mm}
(3) \textit{Filter Significance Identification}: Inside each cluster, based on the gradients analysis, each filter's relative accuracy contribution is further evaluated. Such a filter significance identification will be applied to determine the pruning priority in cluster level pruning;

\vspace{0.5mm}
(4) \textit{Model-wise Filter Pruning}: 
	Given a global pruning ratio $R$,  we multiply it with the layer-wise coefficients to get each layer's actual  pruning ratio $r_l$. The layer-wise coefficients are determined by each layer's pruning accuracy impact as we will show later. 
	For each layer, this layer-wise pruning ratio $r_l$ will be applied to every filter cluster.

\vspace{0.5mm}
(5) \textit{Model Fine-tuning}: After model pruning, a small number of retraining iterations might be applied to recover potential accuracy drop.

\vspace{0.5mm}
The algorithm details are presented as follows:

\noindent\textbf{Filter Functionality Interpretation}
In AM visualization, each filter's functionality is defined as its feature extraction preference from the CNN inputs.
The feature extraction preference of the $i_{th}$ filter $\mathcal{F}_i^l$ in the $l_{th}$ layer is represented by a synthesized input image $X$ that can cause the maximum activation of $\mathcal{F}_i^l$ (\textit{i.e.} the convolutional feature map value).
	The synthesis process of such an input image can be formulated as:
\vspace{-1.5mm}
\begin{small}
\begin{equation}
	V(\mathcal{F}^l_i)=\argmax_{X} {A^l_i(X),
	\hspace{0.6cm} X \leftarrow X} + \eta \cdot \frac{\partial A^l_i(X)}{ \partial X},
	\label{eq:am}
\vspace{-1.5mm}
\end{equation}
\end{small}
where $A^l_i(X)$ is the activation of filter $\mathcal{F}_i^l$ from an input image $X$, $\eta$ is the gradient ascent step size.
	With $X$ initialized as an input image of random noises, each pixel of this input is iteratively changed along the $\partial A^l_i(X)$/$\partial X$ increment direction to achieve the maximum activation.
Eventually, $X$ demonstrates a specific visualized pattern $V(\mathcal{F}^l_i)$, which contains the filter's most sensitive input features with certain semantics, and represents the filter's functional preference for feature extraction.

\begin{figure}%
\centering
\parbox{0.35\textwidth}{
\begin{scriptsize}
\begin{tabular}{p{0.3in}p{0.2in}p{0.2in}p{0.4in}}
		\toprule
		Layer     & K  & Filters &  Ratio \\
		\midrule
		Conv1\_1  &   15     &  62  &  96.8\%     \\
		Conv1\_2  &   17     &  64  &  100\%      \\
		Conv2\_1  &   20     &  128 &  100\%      \\
		Conv2\_2  &   31     &  121 &  94.5\%     \\
		Conv3\_1  &   26     &  251 &  98.0\%     \\
		Conv3\_2  &   24     &  251 &  98.0\%     \\
		Conv3\_3  &   14     &  239 &  93.4\%     \\
		Conv4\_1  &   27     &  447 &  87.3\%     \\
		Conv4\_2  &   28     &  437 &  85.4\%    \\
		Conv4\_3  &   38     &  439 &  85.7\%    \\
		Conv5\_1  &   46     &  500 &  97.7\%    \\
		Conv5\_2  &   61     &  503 &  98.2\%    \\
		Conv5\_3  &   40     &  506 &  98.8\%    \\
		\midrule
		Sum       &   387    & 3942 &  93.3\%   \\
		\bottomrule
	\end{tabular}
\end{scriptsize}
\caption{Filter Cluster Summary}
\label{tab:cluster}
}
\qquad
\begin{minipage}[c]{0.5\textwidth}%
\centering
    \includegraphics[width=1\textwidth]{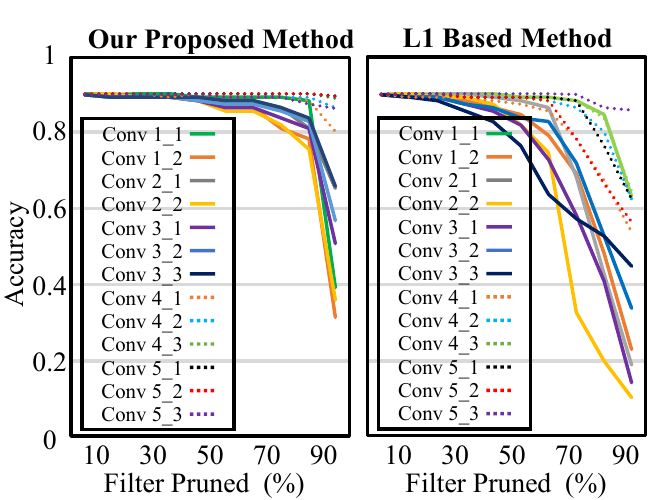}
\vspace{-5mm}
\caption{Individual Layer's Accuracy Impact}
\label{fig:sensi}
\end{minipage}
\vspace{-4mm}
\end{figure}

\vspace{1mm}
\noindent\textbf{Functionality Redundancy Identification}
To identify the functionality redundancy, we apply \textit{k}-means algorithm with pixel-level Euclidean-distance to cluster the filters with similar visualized patterns in each layer.
The pixel-level Euclidean-distance between the AM visualized patterns of filter $\mathcal{F}_k^{l}$ and $\mathcal{F}_i^{l}$ is formulated as:
\vspace{-1.5mm}
\begin{small}
\begin{equation}
	S_{E}[V(\mathcal{F}_{i}^{l}), V(\mathcal{F}_{k}^{l})]=\|V(\mathcal{F}_i^{l}) - V(\mathcal{F}_k^{l}) \|^2.
	\label{eq:sd}
\vspace{-1.5mm}
\end{equation}
\end{small}
which indicates the functionality similarity of any two convolutional filters.

To determine the proper number of clusters (\textit{i.e.} $K$), we perform a grid search from one to half of the total filter number (\textit{i.e.} $I_{l}/2$) in each layer.
	With a larger cluster number, smaller pattern differences are taken into consideration, causing many clusters may only contain one filter with extremely minimal similarity with others.
These filters are considered as non-clustered filters and merged in a locked cluster, which are considered to have unique features and will not be pruned.
	The maximal $K$ during the grid search is selected as final parameter. 

In Fig.~\ref{tab:cluster}, we show the filter cluster distribution based on \textit{k}-means analysis.
	For a VGG-16 model with 13 convolutional layers, the filters in each layer are grouped into 14 to 61 clusters.
	With the layer depth increment, the cluster number also becomes larger: in layer Conv5\_2, the cluster number is as large as 61.
	This is because of the feature complexity increases with more divergent visualized graphic patterns.
Meanwhile, our proposed method can effectively cluster most filters. The minimum cluster ratio across all convolutional layers remains above 85\%.
In average, about 93\% filters are well clustered through the whole model, indicating our method's sufficient filter redundancy analysis capability.

\vspace{1mm}
\noindent\textbf{Filter Significance Identification}
Ideally, the filters with the same functionality can substitute each other.
	However, they may still have slight contribution difference to the prediction accuracy.
	To identify each filter's contribution to the output, we use the first-order Taylor expansion to approximate the CNN output variation under the filter's impact:
	\vspace{-1.5mm}
	\begin{small}
	\begin{equation}
		Z(A_{i}^{l}+\Delta) = Z(A_{i}^{l}) + \frac{\partial Z(A_{i}^{l})}{\partial A_{i}^{l}} \cdot \Delta,~ where ~ \Delta = A_i^l \rightarrow 0,
		\label{eq:1}
		\vspace{-1.5mm}
	\end{equation}
	\end{small}
where $Z(A_{i}^{l}+\Delta)$ is the CNN output loss and the $A_{i}^{l}$ is $i_{th}$ filter’s output feature map in layer $l$.
When filter $\mathcal{F}_{i}^{l-1}$ in the $l-1$ layer is pruned, the filter's output feature map is corresponding set to zero, i.e. changing the $i_{th}$ dimension of $A^l$ to zero.
Therefore, the influence on $Z$ can be qualitatively evaluated by its coefficient $\frac{\partial Z(A_i^l)}{\partial A_i^l}$.


Before pruning, each filter $\mathcal{F}^{(c,l)}_i$ in the cluster $C_l^k$ of layer $l$ is firstly ranked by the contribution index, which is calculated by examining the average gradients:
		\vspace{-1.5mm}
	\begin{small}
	\begin{equation}
		\medmuskip=-2mu
		I(\mathcal{F}^{(c,l)}_i) = \frac{1}{N}\sum_{n=1}^{N} \left \| \frac{\partial Z(F, A^{l})}{\partial A_{i}^{l}(x_{n})} \right \|,
		\label{eq:contri}
		\vspace{-1.5mm}
	\end{equation}
	\end{small}
where $Z(F, A^{l})$ is the CNN output loss of a test image $x_n$, and $A_i^l(x_{n})$ is the feature map of filter $\mathcal{F}^{(c,l)}_i$ for each test image $x_n$.

\begin{figure}[t]
	\centering
	\includegraphics[width=5in]{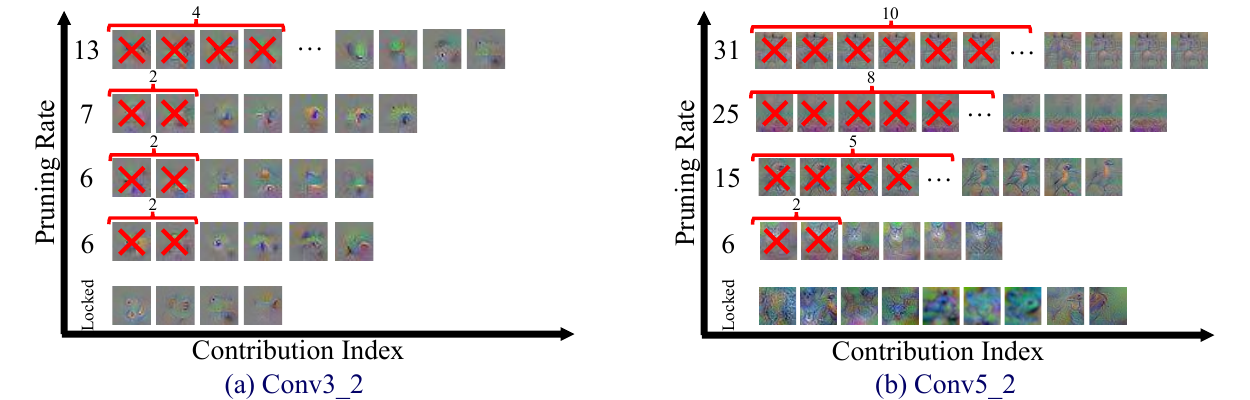}
	\vspace{-1.5mm}
	\caption{Case study of the filter functionality-oriented filter pruning on the Conv3\_2 and Conv5\_2 of VGG-16.
		The convolutional filters are shown by their visualized patterns, and aligned according to the contribution index increment.}
	\label{fig:pruning}
	\vspace{-4mm}
\end{figure}

\vspace{1mm}
\noindent\textbf{Filter Pruning and Fine-Tuning}
Based on the functionality redundancy and filter significance identification, we can proceed to the stage of model-wise convolutional filter pruning.
	Given a global model pruning ratio $R$, the layer-wise pruning ratio of each layer $r_{l}$ is determined by multiplying $R$ and the layer-wise coefficients, which are determined according to each layer's pruning accuracy impact. 
	For example, Fig.\ref{fig:sensi} shows the layer-wise network accuracy impact according to our method and $\ell_1$-norm based method.
	Clearly, the shallow layers demonstrate higher accuracy impact, while the deeper layers have lower impact. 
Therefore, we prune more gently for shallower layers but more aggressive for deeper layers.
The layer-wise coefficients of the pruning ratios $r_{l}$ from the shallow layers to deep layers (conv1\_x to conv5\_x) is thus set to 0.25:0.125:0.125:0.375:0.375.
For other models, dedicated pruning ratios are discussed in later experiments.

Meanwhile, we can see that our proposed pruning has a slower accuracy degradation rate compared with the $\ell_1$-norm based pruning method in majority of layers. This means the redundant filters can be more accurately identified and pruned through the proposed interpretive functionality-oriented filter pruning approach.

Fig.~\ref{fig:pruning} shows our functionality-oriented filter pruning examples with the convolutional layers of Conv3\_2 (a) and Conv5\_2 (b) with $r_l=$ 30\%.
The convolutional filters are shown by their visualized functionality patterns, and the filters with similar patterns are grouped into multiple clusters. 
	When the layer-wise pruning ratio is assigned, the filters with small contribution index values will be firstly pruned.
	According to our proposed method, larger clusters are pruned with more filters. This make senses since such balanced functionality reduction will lead to non-biased functionality composition as the original model's. 
	Therefore, we can see different amounts of filters are pruned between different clusters but all the pruning ratios are $\sim$30\% as shown in the Fig.~\ref{fig:pruning}.

Leveraging the convolutional filters' qualitative functionality analysis, our proposed method is expected to leverage the neural network interpretability for more accurate redundant filter allocation, faster pruning speed, and optimal computation efficiency.

After the model-wise pruning, a small amount of model fine-tuning will be applied to recover the potential accuracy drop, as we will show in later experiments.
\section{Experiments}
We evaluate our proposed method with both single-branch CNN models (\textit{i.e.}, ConvNet, VGG~\cite{Simo:2014:arXiv}) and a multiple-branch CNN model (\textit{i.e.}, ResNet~\cite{He:2016:CVPR}) on CIFAR10/100~\cite{cifar10} and a subset of ImageNet~\cite{imagenet}.
	A data argumentation procedure is applied to CIFAR10/100 dataset through the horizontal flip and random crop, generating a 4-pixel padded training dataset.
	The visualization analysis and filter pruning are implemented in Caffe environment~\cite{jia2014caffe}.
	The contribution index is calculated by using the whole training dataset for more accurate CNN model analysis.
	The retraining process for our method is executed only by 40 epochs (\textit{i.e.}, 1/4 of the original training epochs) with a constant learning rate of 0.001.
	While the retraining process for the compared state-of-the-art will be continuously applied until model convergence.
Specifically, the compared works include $\ell_1$-norm Pruning~\cite{Li:2016:pruning}, Taylor Pruning~\cite{molchanov2016taylor}, Geometric Median Pruning (GM)~\cite{he2018GM}, and Channel Pruning~\cite{he2017channel}.




\vspace{-2mm}
\subsection{CIFAR Experiment}
\vspace{-2mm}
In this section, we first evaluate our method on two CIFAR image datasets, namely CIFAR-10 and CIFAR-100.
The experimental results are shown in Table~\ref{tab:CIFAR}.
\begin{table}[t]
\small
\caption{CIFAR Pruning Comparison}
\centering
\begin{threeparttable}
\begin{tabular}{llllllll}
\toprule
       CNN&    Pruning     &    Baseline   & FLOPs        & FLOPs              & Prune      & Retrain    \\
    Models&Methods   &    acc. (\%)  & (x$10^{8}$)  & $\downarrow$ (\%)  & acc. (\%)  & acc. (\%)  \\ \bottomrule

ConvNet   &$\ell_1$-norm*  &  90.05 &  5.34  &  37.4 & 83.29          &  88.53         \\
(CIFAR10) &Taylor*         &  90.05 &  5.34  &  37.4 & 85.29          &  89.42         \\
          &Ours~(40\%)           &  90.05 &  5.34  &  37.4 & \textbf{87.88} &  \textbf{90.04}\\\bottomrule

VGG-16    &$\ell_1$-norm~\cite{Li:2016:pruning}&        93.25 &  2.06      &  34.2          & -              &  93.40         \\
(CIFAR10) &Taylor*           &        93.25 &  1.85      &  44.10         & 73.24          &  92.31         \\
          &GM~\cite{he2018GM}      &        93.58 &  2.01      &  35.9          & 80.38          &  \textbf{94.00}\\
          &Ours~(45\%)        &        93.25 &  1.85      &  \textbf{44.10}& \textbf{91.13} &  93.30         \\\bottomrule

ResNet-56 &$\ell_1$-norm~\cite{Li:2016:pruning}  &        93.04 &  0.91      &  27.60          & -     &  93.06         \\
(CIFAR10) &Taylor*           &        92.85 &  0.71      &  43.00          & 76.32 &  92.01         \\
          &Channel~\cite{he2017channel}      &        92.80 &  0.56      &  \textbf{50.00} & -     &  93.23         \\
          &Ours~(40\%)        &        92.85 &  0.71      &  43.00          & 81.13 &  \textbf{93.30} \\\bottomrule

VGG-16    &$\ell_1$-norm*     &   73.14 &  1.96      &  37.32       & 63.21           & 72.31         \\
(CIFAR100)&Taylor*            &   73.14 &  1.96      &  37.32       & 65.19           & 72.52         \\
          &Ours~(45\%)         &   73.14 &  1.96      &  37.32       & \textbf{68.21}  & \textbf{73.21}\\\bottomrule
\end{tabular}
\label{tab:CIFAR}
 \begin{tablenotes}
        \footnotesize
        \item ``*'' indicates our implementation. Ours(40\%) means 40\% filters are pruned by our method.
      \end{tablenotes}
    \end{threeparttable}
\vspace{-4mm}
\end{table}

\vspace{1mm}
\noindent\textbf{ConvNet on CIFAR-10}
Our ConvNet is designed based on the AlexNet model with 5 convolutional layers, which contain 96-256-384-384-256 filters respectively.
	All the convolutional filters are constructed with a 3x3 kernel size.
	The baseline test accuracy of our ConvNet is 90.05\%.
As shown in the Table~\ref{tab:CIFAR}, ours method pruned 40\% convolutional filters without accuracy drop and achieved 37.4\% computation load reduction
	We also implemented the $\ell_1$-norm and Taylor pruning for comparison.
	These methods pruned the same amount of filters as our method to achieve the same amount of FLOPs reduction.
	As shown in the Table~\ref{tab:CIFAR}, our method achieves better performance in terms of both pruned accuracy and retrained accuracy.

\vspace{1mm}
\noindent\textbf{VGG-16 on CIFAR-10}
	We implemented a VGG-16 model following the same pre-processing and hyper-parameters configurations as~\cite{Li:2016:pruning}, which consists of 13 convolutional layers and 2 fully-connected layers.
	Each convolutional layer is followed by a batch normalization layer.
As shown in Table~\ref{tab:CIFAR}, our method has better performance than the $\ell_1$-norm and Taylor pruning.
	Under a pruning ratio of 45\%, our proposed method can achieve 44.1\% FLOPs reduction with the accuracy well-retained.
	Although the GM pruning achieves an even improved accuracy performance after the retraining process, it only reduced 35.9\% computation load, about 18.6\% less than our proposed method.

\vspace{1mm}
\noindent\textbf{ResNet-56 on CIFAR-10}
ResNet-56 is a multiple-branch CNN model, which contains three convolutional stages of residual blocks connected by projection mapping channels, one global average pooling layer, and one fully-connected layer.
	After trained on CIFAR-10 from scratch using the same training parameters as~\cite{He:2016:CVPR}, the model can achieve a baseline accuracy of 92.85\%.
	To avoid changing the input and output feature maps of each residual block, same as $\ell_1$, we only prune filters from the first layers of each block.
	Considering ResNet-56 model has many layers, we set the pruning ratio of all convolutional layers to the same value in this experiment.
Our method demonstrate better accuracy retaining capability compared to the $\ell_1$-norm and Taylor pruning.
	Comparing to the Channel Pruning, although our computation load reduction is slightly smaller, our implementation is much more straightforward, considering the Channel Pruning requires additional multiple-branch enhancement to generalize itself to ResNet.

\vspace{1mm}
\noindent\textbf{VGG-16 on CIFAR-100}
We further evaluated our pruning method on the CIFAR100 dataset.
Using the same VGG-16 model, our method introduces less accuracy drop compared to state-of-the-art with the same computation load reduction.
Meanwhile, the accuracy drop caused by our pruning method can be fully recovered with 0.07\% accuracy improvement.

\vspace{-2mm}
\subsection{ImageNet Experiment}
\vspace{-2mm}
In evaluations on ImageNet, 10 and 100 classes of images out of 1000 classes, namely ImageNet-10 and ImageNet-100 are utilized in this work.
Each class contains 1300 training images and 50 validation images.
We evaluate our proposed method for the VGG-16, and ResNet-32 models on the ImageNet-10 and ImageNet-100 respectively.

The VGG-16 model is implemented as the same architecture as the VGG-16 model on CIFAR.
The proportion of the pruning ratios from shallow layers to deep layers (conv1\_x to conv5\_x) is set to 2:1:1:3:3.
The ResNets-32 model for ImageNet have three stages of residual blocks, which contain 3, 4, and 2 residual blocks respectively, and each residual block has three convolutional layers.
Same as ResNet-56 model, in each residual block, only the first two convolutional layers are pruned to keep the input and output feature maps to be identical.
The pruning ratio of all convolutional layers is set to the same value for simplicity.
The result of ImageNet pruning comparison is shown in the Table~\ref{tab:imagenet}.

\noindent\textbf{ImageNet-10}
Table~\ref{tab:imagenet} shows that our method outperforms previous methods on ImageNet10 dataset.
For the VGG-16 model, large accuracy loss occurs in the three pruning method.
However, our proposed method induces less accuracy degradation compared with the $\ell_1$ and Taylor.
With re-training, our method achieves higher accuracy than the baseline accuracy.
Using the ResNet-32, it is hard to recover the accuracy drop through retraining on the ImageNet-10 dataset, and hence acceptable pruning rate in this scenario is relatively small.
As shown in the Table~\ref{tab:imagenet}, we can achieve 32.90\% FLOPs reduction on ResNet-32 with only 0.06\% accuracy drop.

\noindent\textbf{ImageNet-100}
With larger image data complexity and more class composition complexity, the feature extraction becomes more complex.
It's become more challenge for the filter functionality identification and clustering.
Our method can still outperform previous methods on ImageNet100 dataset.
As shown in the Table~\ref{tab:imagenet}, we can achieve 50.64\% FLOPs reduction on VGG-16 with 1\% accuracy improvement.
For the ResNet-32, our method can still achieve less accuracy and better retraining accuracy improvement than the previous method.

\begin{table}[t]
\small
\caption{ImageNet Pruning Comparison}
\centering
\begin{threeparttable}
\begin{tabular}{lllllll}
\toprule
  CNN          &Pruning &    Baseline  & FLOPs        & FLOPs            & Prune      & Retrain   \\
  Models       &Methods &    acc. (\%) & (x$10^{10}$) & $\downarrow$ (\%)& acc. (\%)  & acc. (\%) \\ \bottomrule

VGG-16         &$\ell_1$-norm*              &  93.76&  0.57 &  62.99 &  21.22 &  92.26         \\
(ImageNet-10)  &Taylor*                     &  93.76&  0.57 &  62.99 &  24.37 &  93.12         \\
               &Ours(40\%)                  &  93.76&  0.57 &  62.99 &  \textbf{25.13} &  \textbf{94.34}\\ \bottomrule

ResNet-32      &$\ell_1$-norm*              &  88.31&  1.55 &  32.90 &  80.37 &  84.75         \\
(ImageNet-10)  &Taylor*                     &  88.31&  1.55 &  32.90 &  81.03 &  85.12         \\
               &Ours(30\%)                  &  88.31&  1.55 &  32.90 &  \textbf{81.25} &  \textbf{88.25}\\ \bottomrule

VGG-16         &$\ell_1$-norm*              &  78.51&  0.76 &  50.64 &  36.35 &   76.56        \\
(ImageNet-100) &Taylor*                     &  78.51&  0.76 &  50.64 &  40.28 &   77.32        \\
               &Ours(30\%)                  &  78.51&  0.76 &  50.64 &  \textbf{43.25} &  \textbf{79.51}\\\bottomrule

ResNet-32      &$\ell_1$-norm*              &  75.34&  1.76 &  23.81  &  63.25 &   72.52         \\
(ImageNet-100) &Taylor*                     &  75.34&  1.76 &  23.81  &  68.14 &   73.43         \\
               &Ours(20\%)                  &  75.34&  1.76 &  23.81  &  \textbf{70.21} &   \textbf{75.40}\\ \bottomrule
\end{tabular}
\label{tab:imagenet}
 \begin{tablenotes}
        \footnotesize
	\item ``*'' indicates our implementation. Ours(40\%) means 40\% filters are pruned by our method.
			\vspace{-5mm}
      \end{tablenotes}
    \end{threeparttable}
\end{table}

\vspace{-2mm}
\subsection{CNN Model Retraining Analysis}
\vspace{-2mm}

In most filter pruning works, the retraining process is essential to compensate the accuracy drop. 
However, as aforementioned, its role still lacks certain research.
	In this work, we also analysis the retraining process quantitatively and qualitatively in filter pruning.

\begin{figure}[t]
  \centering
  \includegraphics[width=5in]{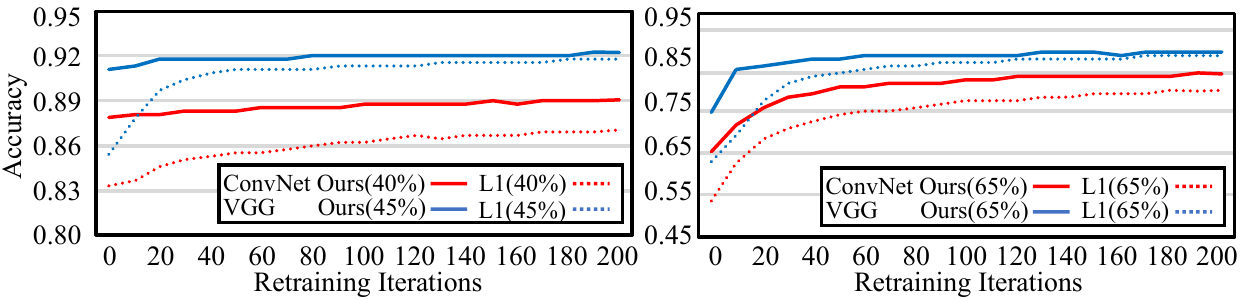}
  \vspace{-2mm}
  \caption{Pruned Model Accuracy Recovery by Retraining.}
  \label{fig:Retrain}
  \vspace{-2mm}
\end{figure}

Fig.~\ref{fig:Retrain} compares the retraining processes of our method and the $\ell_1$-norm based pruning method with ConvNet and VGG-16 on CIFAR-10.
	Comparing our method (solid line) with the $\ell_1$-norm based method (dashed line), we can observe that:
	The models pruned by our method always demonstrate quicker accuracy recovery.
	Taking VGG-16 as an example, our method is close to the convergence accuracy after only after 100 iterations with less retraining dependency.
	While the accuracy of the $\ell_1 $-norm based method is still $\sim$3\% lower than ours, resulting in significantly more retraining effort.

\begin{figure}[t]
  \centering
  \includegraphics[width=5in]{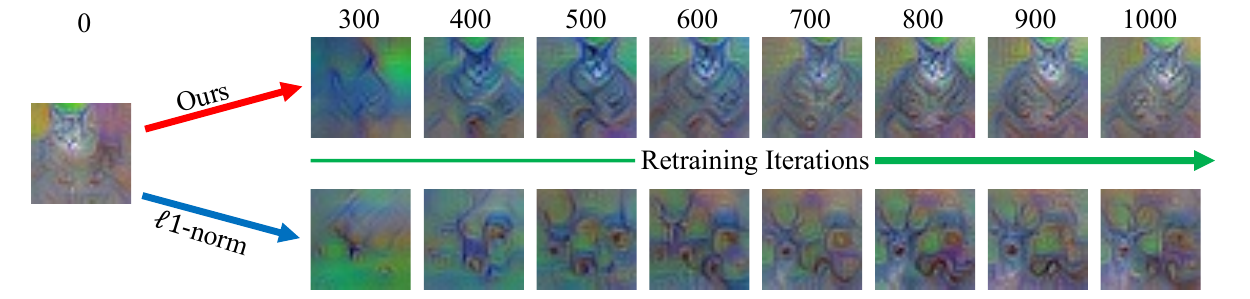}
    \vspace{-3mm}
  \caption{Filter Functionality Transformation During Retraining.}
  \label{fig:Retrain_pattern}
  \vspace{-3mm}
\end{figure}

We also use AM visualization to analysis the filter functionality transition during the retraining process as shown in Fig.~\ref{fig:Retrain_pattern}.
	We randomly choose and visualize one preserved filter after pruning with our method and the $\ell_1$-norm based method.
	During the retraining process, we visualize its functionality pattern every $100$ iterations.
	From Fig.~\ref{fig:Retrain_pattern}, we can see that, the visualized pattern after the $\ell_1$-norm pruning demonstrates dramatic change.
	This indicates that the $\ell_1$-norm based method significantly defects the original CNN model's functionality integrity, and the retraining process has to reconfigure the filter's functionality for compensation.
	Meanwhile, the filter functionality pattern remains unchanged in our method, which indicates our method's precise redundant filter identification.

\vspace{-2mm}
\section{Conclusion}
\vspace{-2mm}
In this work, through convolutional filter AM visualization and functionality analysis, we firstly demonstrate that filter redundancy exists in the form of functionality repetition.
Then, we shows that such functional repetitive filters could be effectively pruned from CNNs to provide computation redundancy reduction. 
Based on such motivation, we propose an interpretable functionality-oriented filter pruning method: By first interpreting and clustering filters with same functions together, we remove the repetitive filters with smallest significance and contribution in a balanced manner inside each cluster. 
Therefore, this implicitly helps maintain the similar functionality composition as the original model, and thus brings less damage to the model accuracy.
Extensive experiments on CIFAR and ImageNet demonstrate the superior performance of our pruning method over state-of-the-art methods.
By analyzing the functionality changing of remaining filters in the retraining process, we further prove our assumption that $\ell_1$-norm based pruning partially destructs original CNNs' functionality integrity.
By contrast, our method shows consistent filter functionality during retraining process, demonstrating less harm to original model functionality.

\section*{Acknowledgments}
This work was supported in part by NSF CNS1717775.

\bibliography{./egbib}
\end{document}